\documentclass[letterpaper, 10 pt, conference]{ieeeconf}  

\IEEEoverridecommandlockouts                              

\usepackage{multirow}
\usepackage{times}
\usepackage{epsfig}
\usepackage{graphicx}
\usepackage{amsmath}
\usepackage{amssymb}
\usepackage{layouts} 
\usepackage{color}
\usepackage{subfig}

\usepackage[breaklinks=true,bookmarks=false]{hyperref}

\title{\LARGE \bf 
End-to-End Race Driving with Deep Reinforcement Learning
}

\author{Maximilian Jaritz$^{1,2}$, Raoul de Charette$^{1}$, Marin Toromanoff$^{2}$, Etienne Perot$^{2}$ and Fawzi Nashashibi$^{1}$
\thanks{$^{1}$Inria, RITS Team, 2 rue Simone Iff, 75012 Paris
        {\tt\small surname.last-name@inria.fr}}%
\thanks{$^{2}$Valeo Driving Assistance Research, Bobigny
        {\tt\small first.lastname@valeo.com}}%
}

\begin{document}

\maketitle
\thispagestyle{empty}
\pagestyle{empty}

\begin{abstract}
We present research using the latest reinforcement learning algorithm for end-to-end driving without any mediated perception (object recognition, scene understanding). The newly proposed reward and learning strategies lead together to faster convergence and more robust driving using only RGB image from a forward facing camera. An Asynchronous Actor Critic (A3C) framework is used to learn the car control in a physically and graphically realistic rally game, with the agents evolving simultaneously on tracks with a variety of road structures (turns, hills), graphics (seasons, location) and physics (road adherence). A thorough evaluation is conducted and generalization is proven on unseen tracks and using legal speed limits. Open loop tests on real sequences of images show some domain adaption capability of our method.
\end{abstract}

\section{Introduction}
Recent advances prove the feasibility of end-to-end robot control by replacing the classic chain of perception, planning and control with a neural network that directly maps sensor input to control output~\cite{levine2016end}. For cars, direct perception~\cite{chen2015deepdriving} and end-to-end control~\cite{Mnih2016AsynchronousLearning} were showcased in the TORCS car racing game using Reinforcement Learning (RL). As RL relies on try and error strategies an end-to-end driving prototype still seems too dangerous for real-life learning and there is still a lot of progress to be done as the first studies use simulators with simplified graphics and physics, and the obtained driving results lack realism.

\begin{figure}
\centering
\subfloat[Overall pipeline (red = for training only)]{\includegraphics[width=0.9\columnwidth]{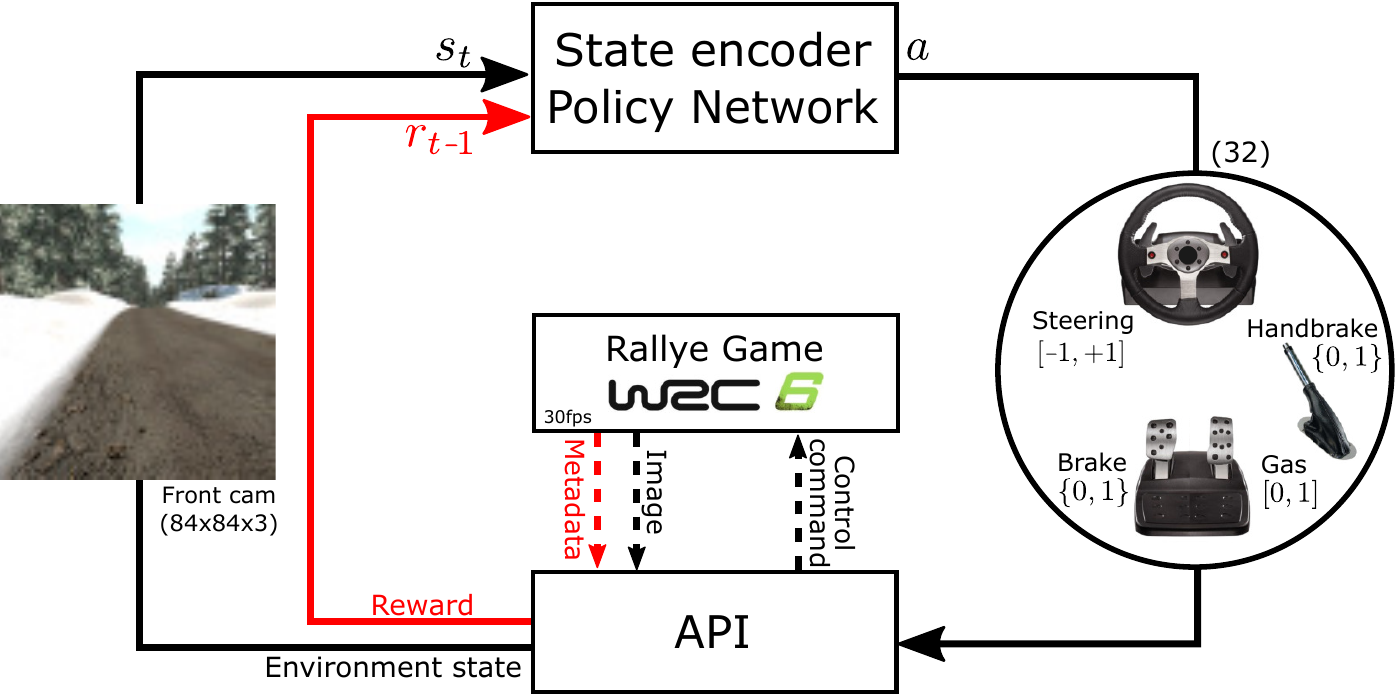}\label{fig:overallPipeline}}\\
\subfloat[Training performance (3 tracks, 29.6km)]{\includegraphics[height=0.26\columnwidth]{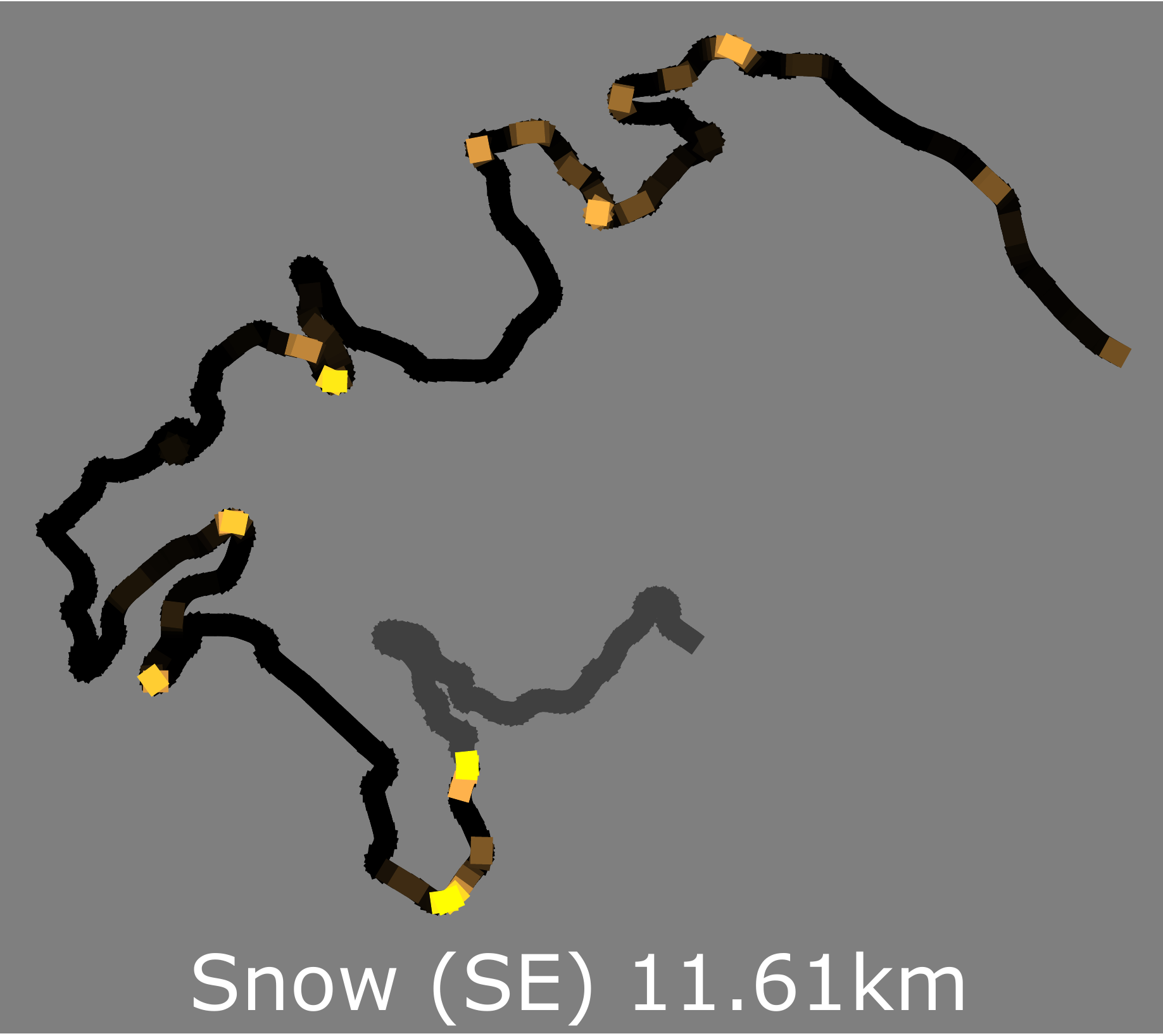}\includegraphics[height=0.26\columnwidth]{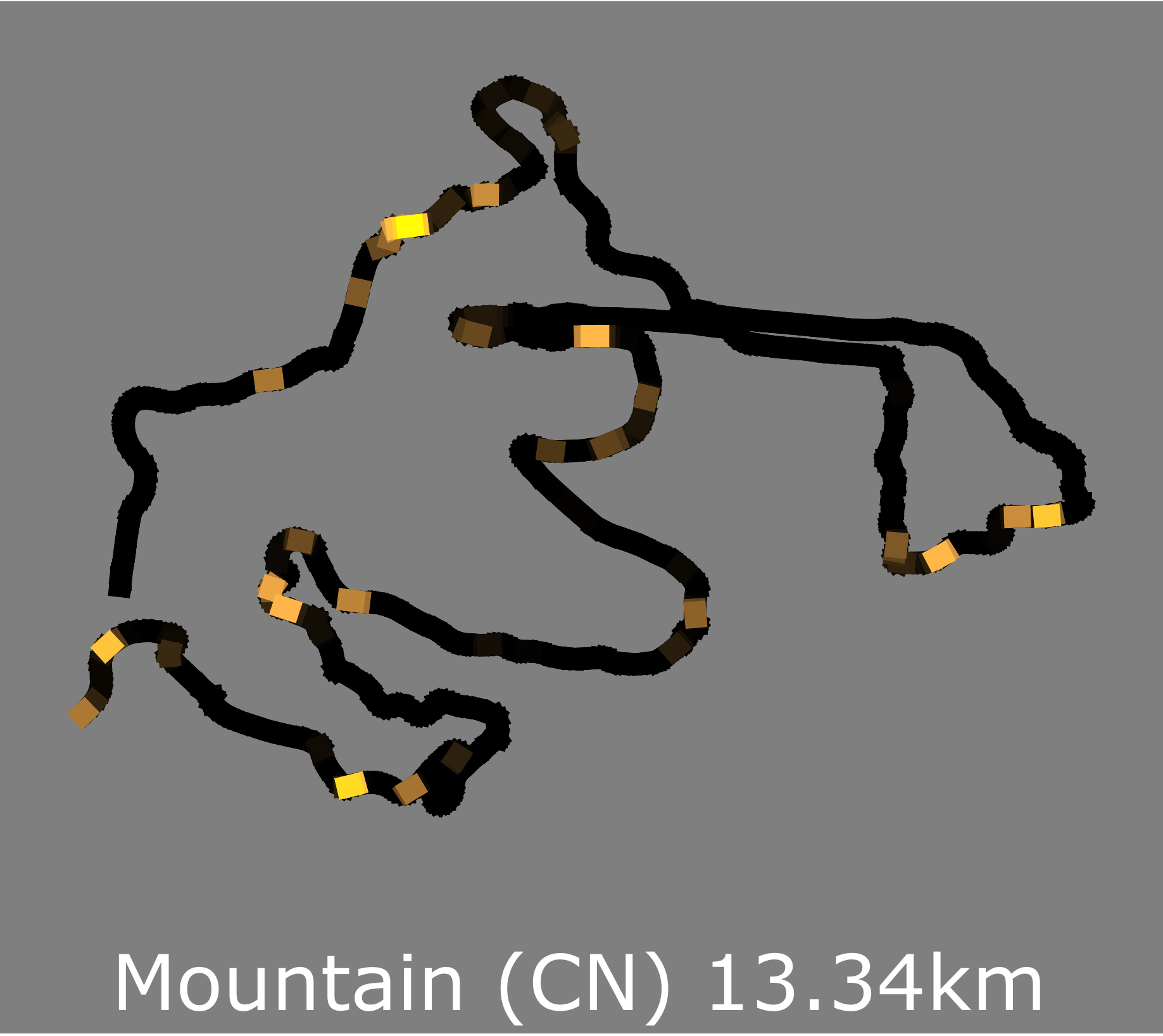}\includegraphics[height=0.26\columnwidth]{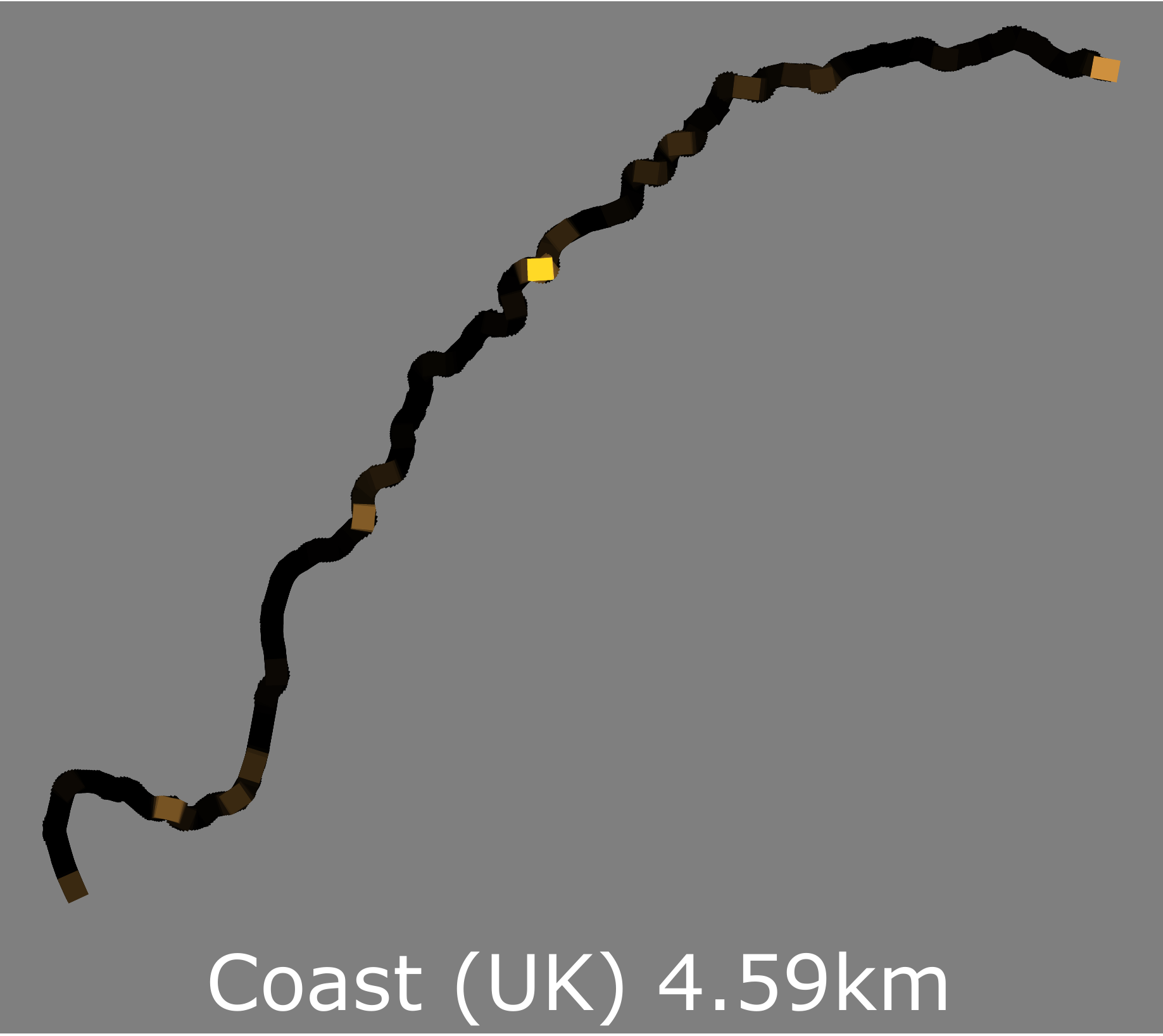}\includegraphics[height=0.26\columnwidth]{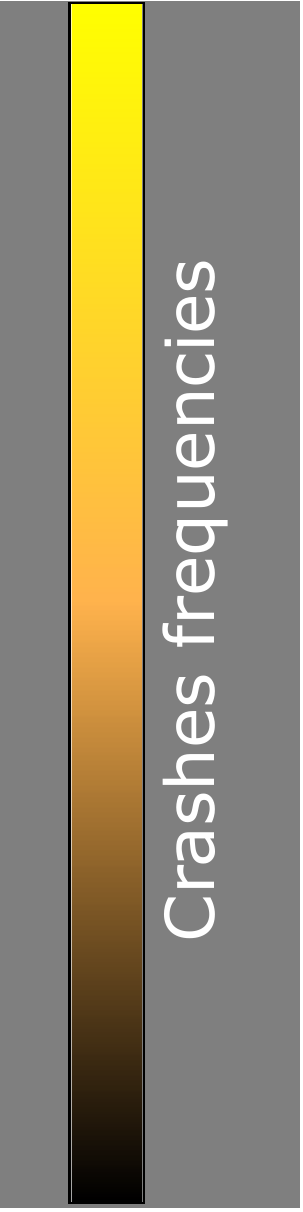}\label{fig:game_graphicsTracks}}
\caption{(a) Overview of our end-to-end driving in the WRC6 rally environment. The state encoder learns the optimal control commands (steering, brake, gas, hand brake), using only 84x84 front view images and speed. The stochastic game environment is complex with realistic physics and graphics. (b) Performance on 29.6km of training tracks exhibiting a wide variety of appearance, physics, and road layout. The agent learned to drive at 73km/h and to take sharp turns and hairpin bends with few crashes (drawn in yellow).}
\label{fig:overall}
\end{figure}

We propose a method (fig. \ref{fig:overall}) benefiting from recent asynchronous learning~\cite{Mnih2016AsynchronousLearning} and building on our preliminary work~\cite{perot2017end} to train an end-to-end agent in World Rally Championship 6 (WRC6), a realistic car racing game with stochastic behavior (animations, light). In addition to remain close to real driving conditions we rely only on image and speed to predict the full longitudinal and lateral control of the car.
Together with our learning strategy, the method converges faster than previous ones and exhibits some generalization capacity despite the significantly more complex environment that exhibits 29.6km of training tracks with various visual appearances (snow, mountain, coast) and physics (road adherence). Although it is fully trained in a simulation environment the algorithm was tested successfully on real driving videos, and handled scenarios unseen in the training (e.g. oncoming cars).

Section~\ref{sec:relatedwork} describes the few related works in end-to-end driving. Section~\ref{sec:method} details our methodology and learning strategies. Exhaustive evaluation is discussed in~\ref{sec:results} and generalization on real videos is shown in~\ref{sec:expGeneralization}. 


\section{Related Work}
\label{sec:relatedwork}
Despite early interest for end-to-end driving~\cite{pomerleau1989alvinn,bajracharya2009autonomous} the trend for self-driving cars is still to use perception-planning-control paradigm~\cite{sun2006road,urmson2008autonomous,montemerlo2008junior}. The slow development of end-to-end driving can be explained both due to the computational and algorithmic limitations that evolved recently with deep learning. There has been a number of deep learning approaches to solve end-to-end control (aka "behavioral reflex") for games~\cite{mnih2015human,mnih2013playing,Mnih2016AsynchronousLearning} or robots~\cite{lecun2005off,levine2016end} but still very few were applied to end-to-end driving.

Using supervised learning, Bojarski et al.~\cite{bojarski2016end} trained an 8 layer CNN to learn the lateral control from a front view camera, using steering angle from a real driver as a ground truth. It uses 72h of training data, with homographic interpolation of three forward-facing cameras to generate synthetic viewpoints.
Similar results were presented in~\cite{rausch2017learning} with a different architecture. 
Another recent proposal from~\cite{xu2016end} is to use privilege learning and to compute the driving loss through comparison with actual driver decision. An auxiliary pixel-segmentation loss is used which was previously found to be a way to help training convergence. However, in~\cite{xu2016end} it isn't clear whether the network learned to detect the driver action or to predict it. Behavioral cloning is limited by nature as it only mimics the expert driver and thus cannot adapt to unseen situations.

An alternative for unsupervised (or self-supervised) learning, is deep Reinforcement Learning (RL) as it uses reward to train a network to find the most favorable state. The two major interests for RL are that: a) it doesn't require ground truth data and b) the reward can be sparse and delayed which opens new horizons. Indeed, judging a driving control decision at the frame level is complex and deep RL allows rewarding or penalizing after a sequence of decisions. To avoid getting caught in local optima, experience replay memory can be used as in~\cite{mnih2013playing}. 
Another solution is asynchronous learning - coined A3C - as proposed by Mnih et al. and successfully applied for learning ATARI games~\cite{Mnih2016AsynchronousLearning} using score as a reward. A3C achieves experience decorrelation with multiple agents evolving in different environments at the same time.
In~\cite{Mnih2016AsynchronousLearning} they also applied deep RL to the TORCS driving game. A 3 layer CNN is trained with an A3C strategy to learn jointly lateral and longitudinal control from a virtual front camera. The algorithm searches itself for the best driving strategy based on a reward computed from the car's angle and speed.

There has also been a number of close studies in \textit{direct perception}. Rather than learning the control policy, it learns the extraction of high level driving features (distance to ongoing vehicle, road curvature, distance to the border, etc.) which are then used to control the car through simple decision making and control algorithms. This field yields interesting research from~\cite{chen2015deepdriving} and recently~\cite{al2017deep}.

\section{Method}
\label{sec:method} 
In this work we aim at learning end-to-end driving in rally conditions while having a variety of visual appearances and physic models (road adherence, tire friction). This task is very challenging and cannot be conducted in real cars, as we aim to learn full control that is steering, brake, gas and even hand brake to enforce drifting. Instead, the training is done using a dedicated API of a realistic car game (WRC6). This simulation environment allows us to crash the car while ensuring we encounter multiple scenarios. We demonstrate in section \ref{sec:expGeneralization} that the simulation training is transposable to real driving images.

We chose the Asynchronous Learning strategy (A3C)~\cite{Mnih2016AsynchronousLearning} to train the architecture because it is well suited for experience decorrelation. The overall pipeline is depicted in fig.~\ref{fig:overallPipeline}.
At every time-step, the algorithm receives the state of the game ($s$), acts on the car through control command ($a$), and gets a reward ($r$) on next iteration as supervision signal.
The complete architecture optimizes a driving policy to apply to the vehicle using only the RGB front view image.

We first motivate our choice for the reinforcement learning algorithms (sec.~\ref{sec:RL-algorithms}), detail the state encoder architecture (sec.~\ref{sec:state-encoder}) and then describe the strategy applied for the training of the reinforcement learning algorithm~\ref{sec:learning-strategy}.

\subsection{Reinforcement learning algorithms}
\label{sec:RL-algorithms}
In the common RL model an agent interacts with an environment at discrete time steps $t$ by receiving the state $s_t$ on which basis it selects an action $a_t$ as a function of policy $\pi$ with probability $\pi(a_t|s_t)$ and sends it to the environment where $a_t$ is executed and the next state $s_{t+1}$ is reached with associated reward $r_t$. Both, state $s_{t+1}$ and reward $r_t$, are returned to the agent which allows the process to start over.
The discounted reward $R_t = \sum_{k=0}^{\infty} \gamma^k r_{t+k}$ with $\gamma \in [0, 1[$ is to be maximized by the agent. 

\subsubsection{Policy optimization}
\label{sec:rl}
The output probabilities of the control commands (i.e. steering, gas, brake, hand brake) are determined by the control policy $\pi_{\theta}$ parameterized by $\theta$ (e.g. the weights of a neural network) which we seek to optimize by estimating the gradient of the expected return $\mathbb{E}[R_t]$.
To do so, we chose the rather popular REINFORCE method~\cite{williams1992simple} that allows to compute an unbiased estimate of $\nabla_{\theta}\mathbb{E}[R_t]$.

\subsubsection{Asynchronous Advantage Actor Critic (A3C)}
\label{sec:async-actor-critic}
As introduced by Mnih et al.~\cite{Mnih2016AsynchronousLearning}, in A3C the discounted reward $R_t$ is estimated with a value function $V^{\pi_{\theta}}(s) = \mathbb{E} \left[ R_t | s_t=s \right]$ and the remaining rewards can be estimated after some steps as the sum of the above value function and the actual rewards: $\hat{R}_t = \sum_{i=0}^{k-1} \gamma^i r_{t+i} + \gamma^{k} \hat{V}^{\pi_{\theta}}(s_{t + k})$ where $k$ varies between $0$ and $t_{max} = 5$ the sampling update.
The quantity: $\hat{R_t} - \hat{V}^{\pi_{\theta}}(s_t)$ can be seen as the advantage, i.e. whether the actions $a_t, a_{t+1}, ..., a_{t+t_{max}} $ were actually better or worse than expected. This is of high importance as it allows correction when non optimal strategies are encountered.\\
For generality, it is said that the policy $\pi(a_t|s_t; \theta)$ (aka Actor) and the value function $\hat{V}(s_t; {\theta'})$ (aka Critic) are estimated independently with two neural networks; each one with a different gradient loss.
In practice, both networks share all layers but the last fully connected.

In addition to its top performance, the choice of this A3C algorithm is justified because of its ability to train small image encoder CNNs without any need of experience replay for decorrelation. This allows training in different environments simultaneously, which is useful for our particular case as the WRC6 environment is undeterministic.

\subsection{State encoder}
\begin{figure}
	\subfloat[]{\includegraphics[height=0.08\textheight]{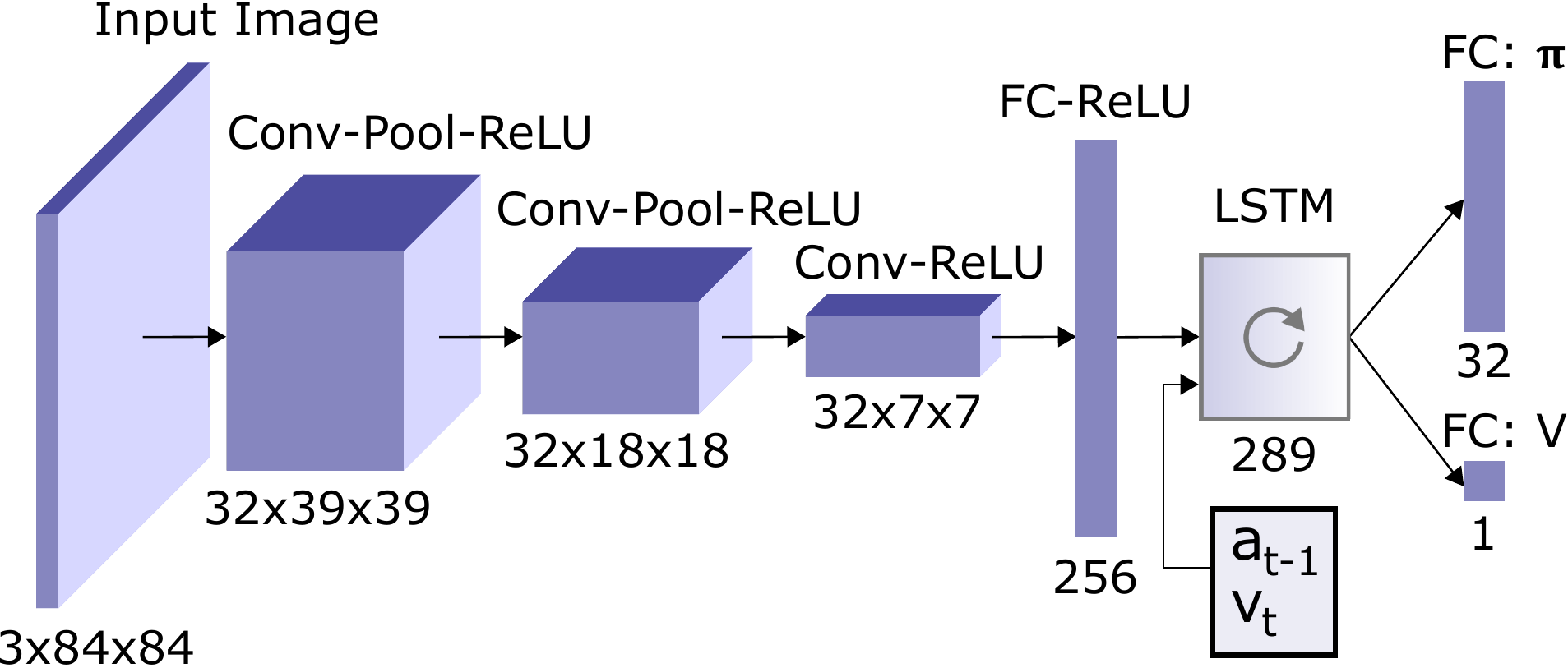}\label{fig:cnnsOurs}}
	\hspace{0.05\columnwidth}
	\subfloat[]{\includegraphics[height=0.08\textheight]{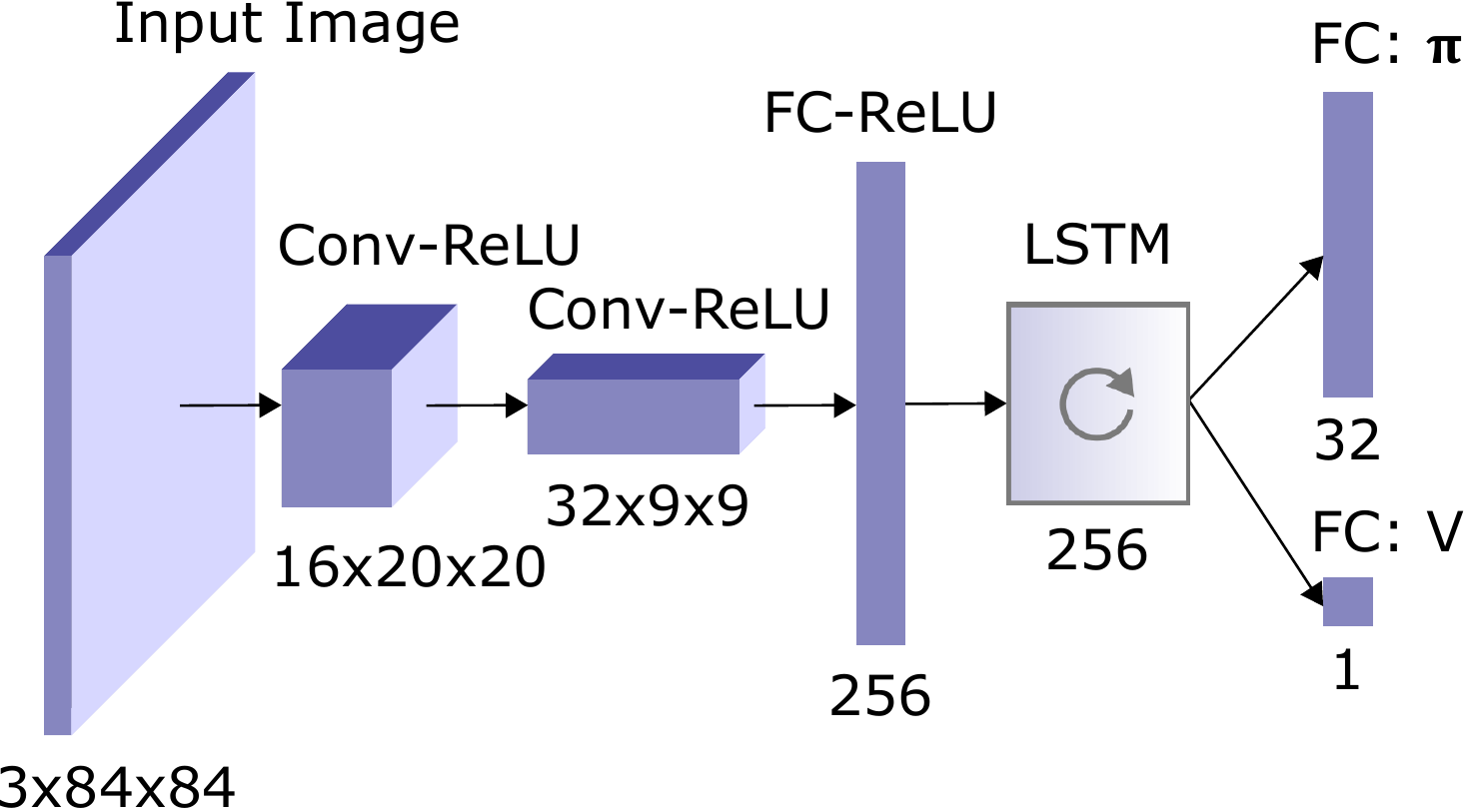}\label{fig:cnnsMnih}}
	\caption{The state encoder CNN+LSTM architecture used in our approach (\ref{fig:cnnsOurs}). Compared to the one used in~\cite{Mnih2016AsynchronousLearning} (\ref{fig:cnnsMnih}), our network is slightly deeper and comes with a dense filtering for finer feature extraction and far-away vision.}
	\label{fig:cnns}
\end{figure}

Intuitively, unlike for other computer vision tasks a shallow CNN should be sufficient as car racing should rely mostly on obstacles and road surface detection. The chosen network is a 4 layer (3 convolutional) architecture inspired by~\cite{kempka2016vizdoom} but using a dense filtering (stride 1) to handle far-away vision. It also uses max pooling for more translational invariance and takes advantage of speed and previous action in the LSTM. Using a recurrent network is required because there are multiple valid control decision if we don't account for motion. Our network is displayed in fig.~\ref{fig:cnns} alongside the one from Mnih et al.~\cite{Mnih2016AsynchronousLearning} which we compare against in section~\ref{sec:results}.
\\


\label{sec:state-encoder}

\subsection{End-to-end learning strategy}
\label{sec:learning-strategy}

Preliminary research highlighted that naively training an A3C algorithm with a given state encoder does not reach optimal performances. In fact, we found that control, reward shaping, and agent initialization are crucial for optimal end-to-end driving. Although the literature somewhat details control and reward shaping, the last one is completely neglected but of high importance.

\subsubsection{Control}
\label{sec:control-strategy}

Continuous control with Deep Reinforcement Learning (DRL) is possible~\cite{lillicrap2015continuous,duan2016benchmarking,pmlr-v48-gu16} but the common strategy for A3C is to use a discrete control, easier to implement. For this rally driving task, the architecture needs to learn the control commands for steering~(-1...1), gas~(0...1), brake~(0,~1) and hand brake~(0,~1). Note that the brake and hand brake commands are binary. The latter was added for the car to learn drifts in hairpin bends, since brake implies slowing down rather than drifting. The combination of all control commands has been broken into 32 classes listed in Table~\ref{tab:controls}. Although the total number of actions is arbitrary, two choices should be highlighted: the prominence of acceleration classes (18 commands with gas~$>$~0) to encourage speeding up, and the fact that brake and hand brake are associated with different steering commands to favor drifting.

\begin{table}[h]
\center
\bgroup
\setlength{\tabcolsep}{0.3em} 
\begin{tabular}{|c|c|c|c|c|}
  \hline
  \multirow{2}{*}{\scriptsize \# classes} 	& \multicolumn{4}{c|}{\scriptsize Control commands} \\
  \cline{2-5} 
  & \scriptsize Steering & \scriptsize Gas & \scriptsize Brake & \scriptsize Hand brake \\
  \hline
  \scriptsize 27 	& \scriptsize $\{-1., -0.75, ..., 1.\}$ 	& \scriptsize $\{0.0, 0.5, 1.0\}$ 	& \scriptsize $\{0\}$ & \scriptsize $\{ 0 \}$ \\
  \scriptsize 4 	& \scriptsize $\{-1., -0.5, 0.5, 1.\}$ 		& \scriptsize $\{0.0\}$ 			& \scriptsize $\{0\}$ & \scriptsize $\{ 1 \}$ \\
  \scriptsize 1 	& \scriptsize $\{0.0\}$ 					& \scriptsize $\{0.0\}$ 			& \scriptsize $\{1\}$ & \scriptsize $\{ 0 \}$ \\
  \hline
\end{tabular}
\egroup
\caption{The 32 classes output for the policy network. First column indicate the number of classes with possible set of control values. Note the prominence of gas commands.}
\label{tab:controls}
\end{table}

The softmax layer of the policy network outputs the probabilities of the 32 control classes given the state encoded by the CNN and LSTM.


\subsubsection{Reward shaping}
\label{sec:reward-shaping}
Reward shaping is crucial to help the network to converge to the optimal set of solutions. Because in car racing the score is measured at the end of the track it is much too sparse to train the agents. Instead, a reward is computed at each frame using metadata information received from the game. 
In~\cite{Mnih2016AsynchronousLearning} the reward is computed as a function of the difference of angle $\alpha$ between the road and car's heading and the speed $v$.
Tests exhibit that that it cannot prevent the car to slide along the guard rail which makes sense since the latter follows the road angle. 
Eventually, we found preferable to add the distance from the middle of the road $d$ as a penalty:

\begin{equation}
R = v (\cos{\alpha} - d)
\label{eq:rewardvcostd}
\end{equation}

Similarly to~\cite{Lau2016UsingTORCS} our conclusion is that the distance penalty enables the agent to rapidly learn how to stay in the middle of the track.
Additionally, we propose two other rewards using the road width as detailed in section \ref{sec:rewardShaping}.


\subsubsection{Agent initialization}

In previous end-to-end DRL attempts~\cite{Mnih2016AsynchronousLearning,Lau2016UsingTORCS} the agents are always initialized at the beginning of the track. Such a strategy will lead to overfitting at the beginning of the training tracks and is intuitively inadequate given the decorrelation property of the A3C algorithm. In practice, restarting at the beginning of the track leads to better performance on the training tracks but poorer generalization capability, which we prove in section \ref{sec:respawnstrat}. 

Instead, we chose to initialize (at start or after crash) the agents randomly on the training tracks although restrained to random checkpoint positions, due to technical limitations. Experiments detailed in section \ref{sec:respawnstrat} advocate that random initialization improves generalization and exploration significantly.

\section{Experiments}
\label{sec:results}
\begin{figure}
	\includegraphics[width=1.0\linewidth]{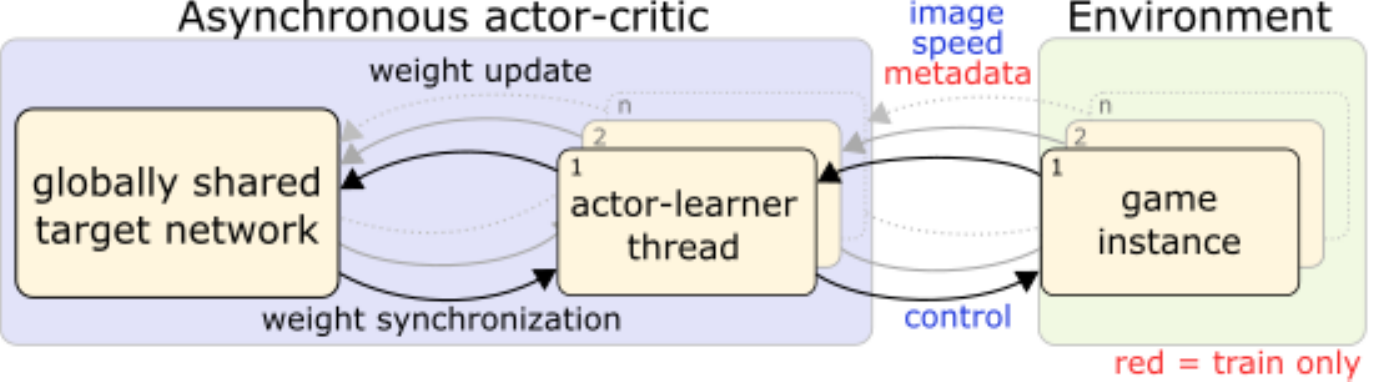}
	\caption{Scheme of our training setup. Multiple game instances run on different machines and communicate through a dedicated API with the actor-learner threads which update and synchronize their weights frequently with the shared target network in an asynchronous fashion.}
	\label{fig:setup}
\end{figure}

This section describes our architecture setup and reports quantitative and qualitative performance in the World Rally Championship 6 (WRC6) racing game. Compared to the TORCS platform used in~\cite{Mnih2016AsynchronousLearning,chen2015deepdriving,Lau2016UsingTORCS}, WRC6 exhibits more realistic physics engine (grip, drift), graphics (illuminations, animations, etc.), and a variety of road shapes (sharp turns, slopes). Additionally, the WRC6 stochastic behavior makes each run unique which is harder but closer to real conditions.

For better graph visualization all plots are rolling mean and deviation over 1000 steps.
Performance is best seen in our video: \url{http://team.inria.fr/rits/drl}.

\subsection{Training setup} 


The training setup is depicted in figure~\ref{fig:setup}. We use a central machine to run the RL algorithm which communicates with 9 instances of the game split over 2 machines. Each of the agents communicates via TCP with a WRC6 instance through a dedicated API specifically developed for this work. 
It allows us to retrieve in-game info, compute the reward and send control back to the game. To speed up the pipeline and match the CNN input resolution, some costly graphics effects were disabled and we used a narrower field of view compared to the in-game view, as shown in figure~\ref{fig:overallPipeline}. 
The game's clock runs at 30FPS and the physical engine is on hold as it waits for the next action. 

We use only first person view images (RGB normalized) and speed for testing to allow a fair comparison with a real driver knowledge. Likewise, training images do not contain usual in-game info (head up display, turns, track progress, etc.) which actually makes the gaming task harder.\\

Three tracks - a total of 29.6km - were used for training (3 instances of each) and we reserved two tracks for testing the generalization of the learning which is discussed in section~\ref{sec:expGeneralization}. The agents (cars) start and respawn (after crashes) at a random checkpoint (always at 0 km/h).

\subsection{Performance evaluation}
\label{sec:performance}

\begin{figure}
\centering
\includegraphics[width=0.99\linewidth]{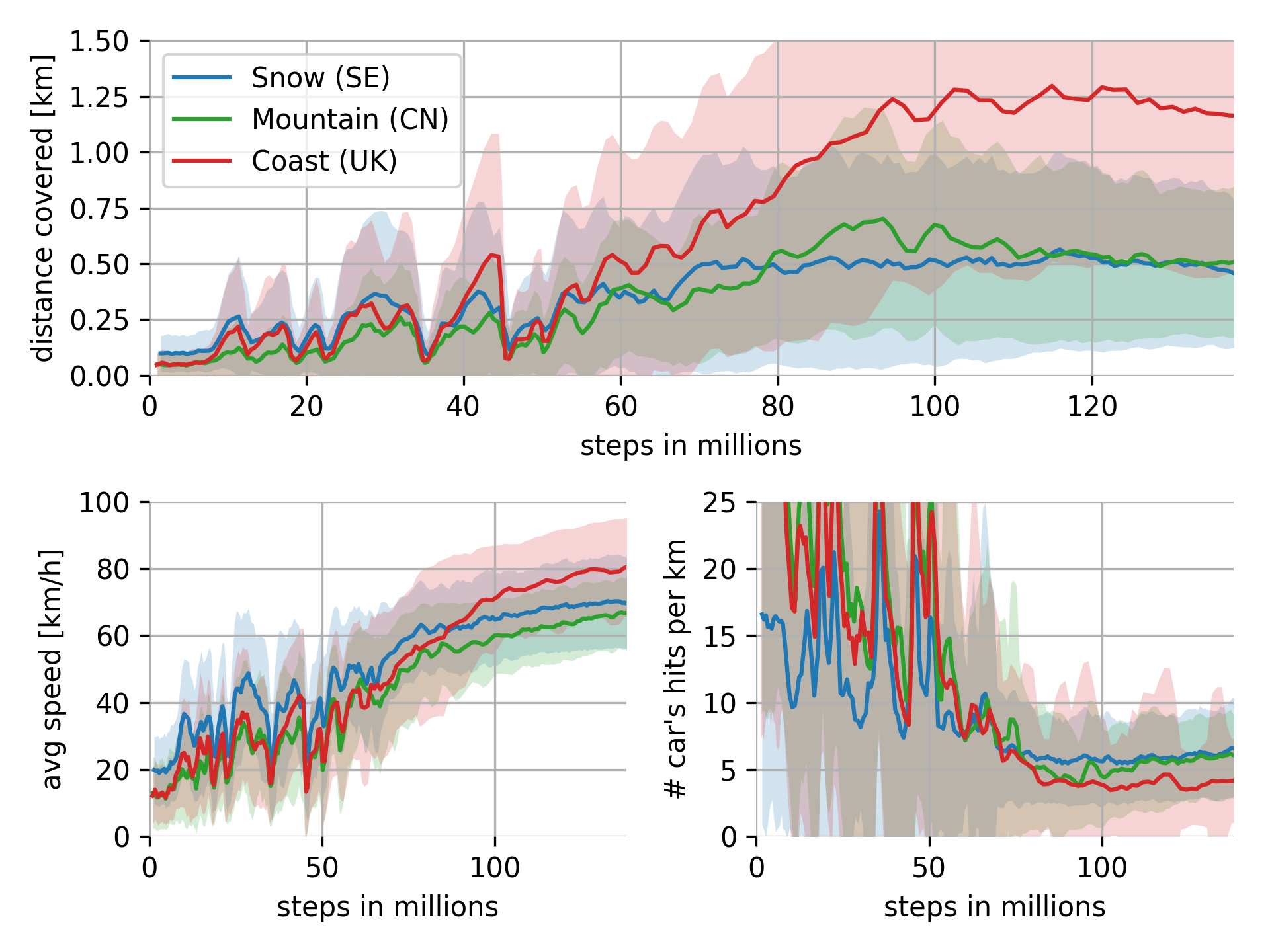}
\caption{Rolling mean (dark) and standard deviation (light) over training on 3 tracks. The agent had more difficulty to progress on the \textit{mountain} and \textit{snow} tracks as they exhibit sharp curves, hairpin bends, and slippery road.}
\label{fig:perfTrainTracks}
\end{figure}

Plots in fig.~\ref{fig:perfTrainTracks} show the mean (dark) and standard deviation (light) over the three training tracks for 140 million steps. During training, the agents progress along the tracks simultaneously as expected from the decorrelation capability of the A3C algorithm. 
Overall, the agents successfully learned to drive despite the challenging tracks appearances and physics at an average speed of 72.88km/h and covers an average distance of 0.72km per run.
We observe a high standard deviation in the covered distance as it depends on the difficulty of the track part where the car is spawned.
The bottom plot reports how often the car hits objects of the environment (guard rail, obstacles, etc.). Although not penalized in the reward function, the hits get lower through training as they imply a speed decrease. After training, the car hits scene objects 5.44 times per kilometer.\\
During training a run is interrupted if the bot either stops progressing or goes in the wrong direction (off road, wrong ways). We refer to this as "crash". 
The location of crashes over five~meter segments is colored from black to yellow in fig.~\ref{fig:game_graphicsTracks}. 
The latter shows the bot learned to go through slopes, sharp curves and even some hairpin bends.

From a qualitative point of view, the agent drives rather smoothly and even learned how to drift with the hand-brake control strategy. 
However, the bots still not achieve optimal trajectories from a racing aspect (e.g. taking turns on the inside). This is explained because the network lacks anticipation, and the car will always try to remain in the track center.
For the snow track the road being very slippery it leads to frequent crashes when the vehicle rushes headlong.
Although on all tracks the average number of hits is relatively high, the context of racing game is very complex and we found during our experiments that even best human players collide with the environment. 
It is important to highlight that the physics, graphics, dynamics, and tracks are much more complex than the usual TORCS platform which explain the lower performance in comparison. In fact, as the architecture learns end-to-end driving it needs to learn the realistic underlying dynamics of the car. Additionally, we trained using tracks with different graphics and physics (e.g. road adherence) thus increasing the complexity.

\begin{figure}
	\centering
	\includegraphics[width=0.23\columnwidth]{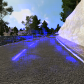}\hspace{0.01\columnwidth}\includegraphics[width=0.23\columnwidth]{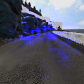}\hspace{0.01\columnwidth}\includegraphics[width=0.23\columnwidth]{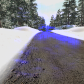}\hspace{0.01\columnwidth}\includegraphics[width=0.23\columnwidth]{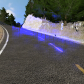}\label{fig:stateencoderVisuBackprop}
	\caption{Visualization of back-propagation where positive gradients for the chosen actions are highlighted in blue. Despite various scenes and road appearances the network learned to detect road edges and relies on them for control.}
	\label{fig:stateencoderVisu}
\end{figure}

To better understand the network decisions, fig. \ref{fig:stateencoderVisu} shows guided back propagation~\cite{Springenberg2014StrivingNet} (i.e. the network positive inner gradients that lead to the chosen action) for several scenarios. Despite the various scene appearances the agent uses the road edges and curvature as a strong action indicator. This mimics existing end-to-end techniques~\cite{Mnih2016AsynchronousLearning,Lau2016UsingTORCS,bojarski2016end} that also learn lateral controls from road gradients. 

\subsection{Comparative evaluation}
To evaluate our contribution as compared to the state of the art we evaluate separately the proposed choices of state encoder, reward shaping and respawn strategies. The study of each factor is carried out by retraining the whole network while only changing the element of study.

\subsubsection{State encoder}
\label{sec:compartiveEvalStateEncoder}
\begin{figure}
\centering
\subfloat[Performance]{\includegraphics[height=0.295\columnwidth]{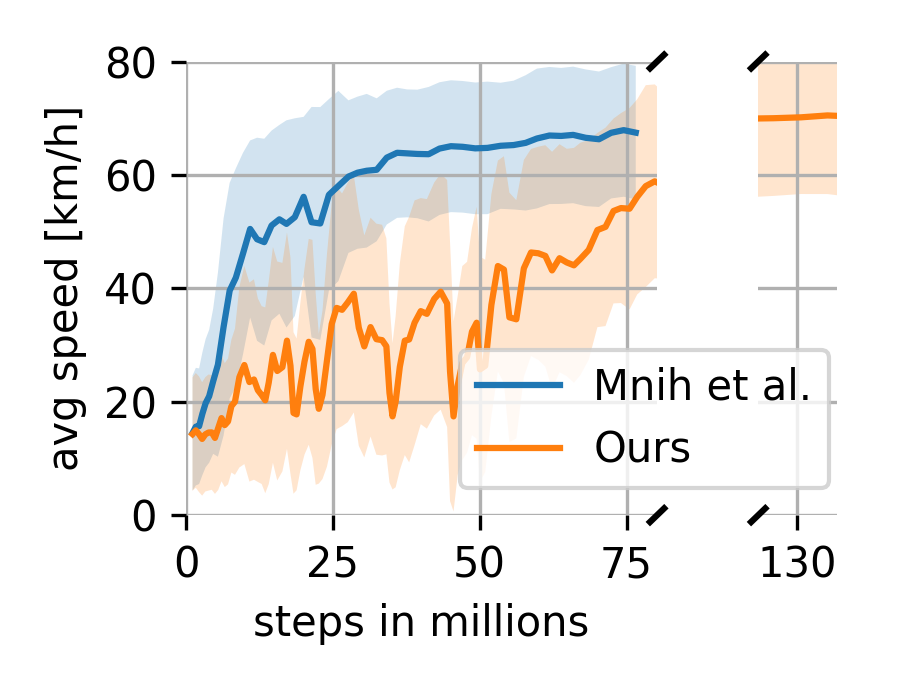}\label{fig:cnncomparisonSpeed}}
\subfloat[Racing style]{\includegraphics[height=0.265\columnwidth]{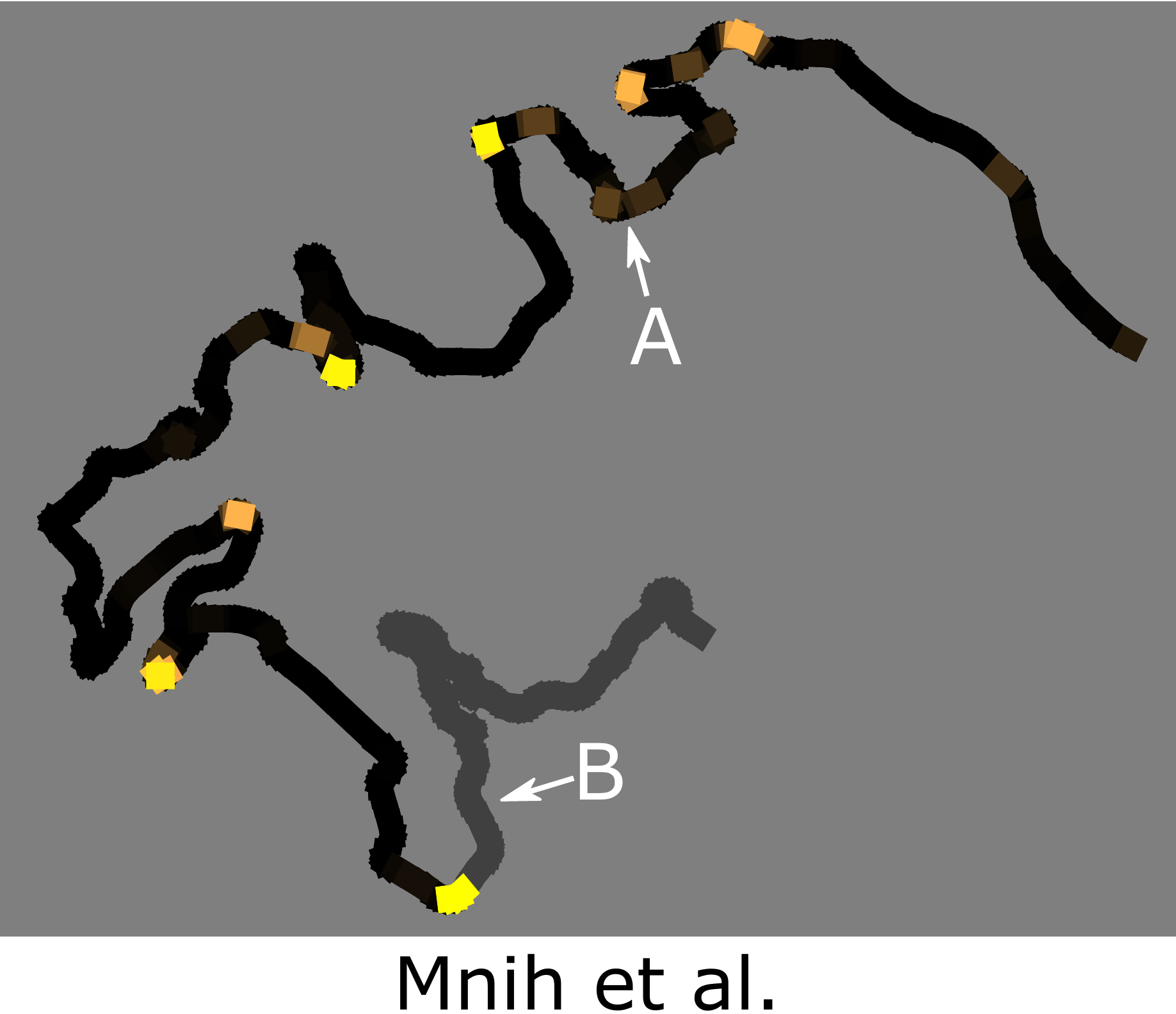}\includegraphics[height=0.265\columnwidth]{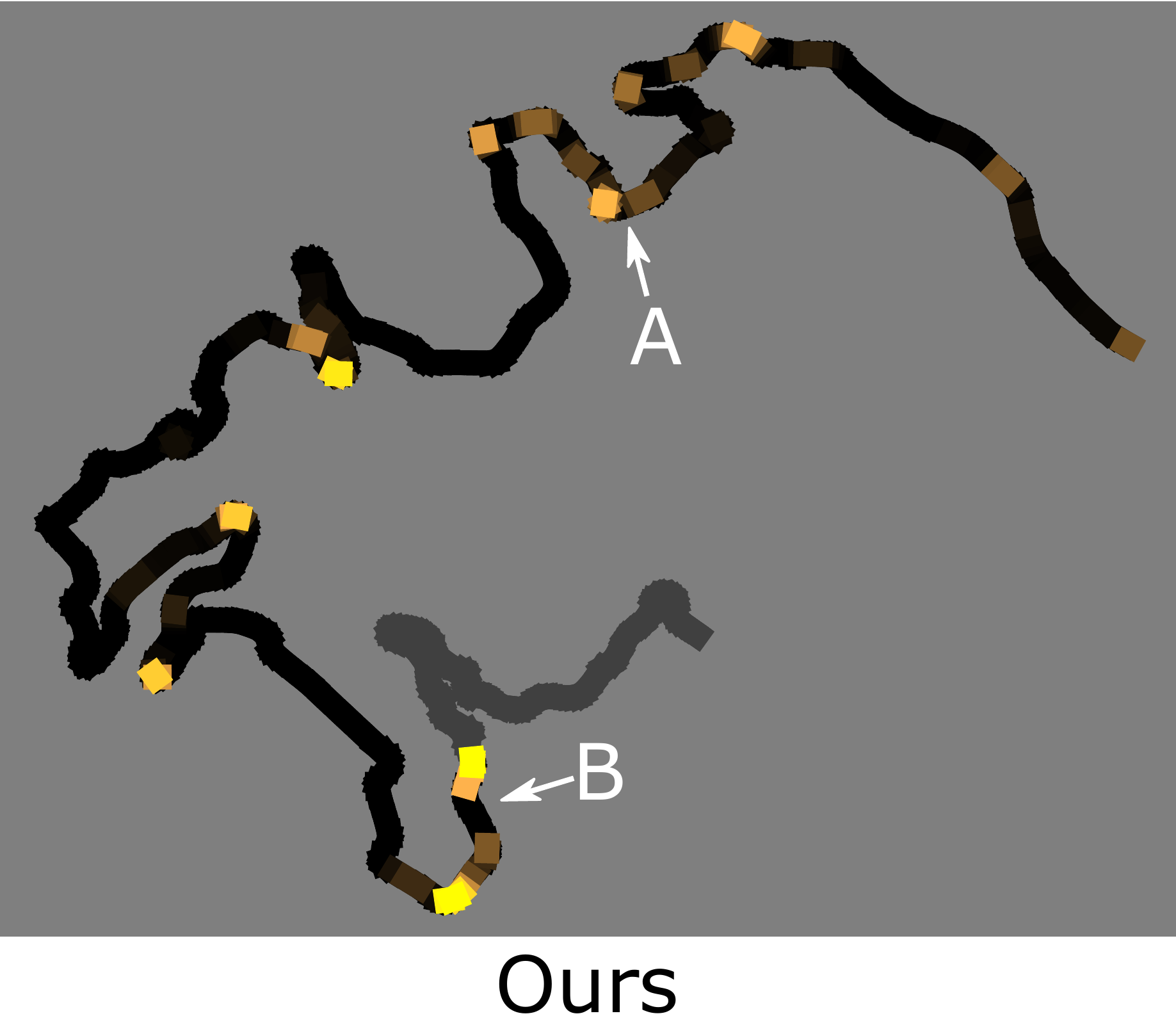}\label{fig:cnncomparisonTrack}}
\caption{Evaluation of our state encoder versus the encoder used in Mnih et al.~\cite{Mnih2016AsynchronousLearning}. In \ref{fig:cnncomparisonSpeed}, the smaller CNN from~\cite{Mnih2016AsynchronousLearning} (cyan) converges faster than ours (orange). In \ref{fig:cnncomparisonTrack} the racing performance of both networks are comparable though slightly more exploratory with our network, as highlighted at location A and B. Refer to section \ref{sec:compartiveEvalStateEncoder} for details.}
\label{fig:cnncomparison}
\end{figure}
Fig.~\ref{fig:cnncomparison} compares the performance of our CNN with dense stride against the smaller network with larger stride from Mnih et al.~\cite{Mnih2016AsynchronousLearning} (cf. fig.~\ref{fig:cnns}). 
As expected, in fig.~\ref{fig:cnncomparisonSpeed} the convergence is faster for the smaller network \cite{Mnih2016AsynchronousLearning} than for our network (80 versus 130 mega steps).
However although a significant impact on the racing style was expected intuitively due to the far away vision, in fig.~\ref{fig:cnncomparisonTrack} both networks seem leading to similar performance.
As a matter of fact, the crashes locations (highlighted in yellow) are similar even on such a difficult track. With the notable exception of the section labeled A where \cite{Mnih2016AsynchronousLearning} crashes less often and section B only explored by our network.\\
In light of these results, we retrained the whole experiment shown in fig. \ref{fig:perfTrainTracks} with the network from \cite{Mnih2016AsynchronousLearning}. Despite the longer convergence, the performance of our network is better compared to the smaller network with $+89.9m$ average distance covered ($+14.3\%$) and $-0.8$ average car's hit per kilometer ($-13.0\%$). Such analysis advocates that our network performs better at a cost of longer training.

\subsubsection{Reward shaping}\label{sec:rewardShaping}
\begin{figure}
	\centering
\subfloat[Reward shaping]{\includegraphics[height=0.28\linewidth]{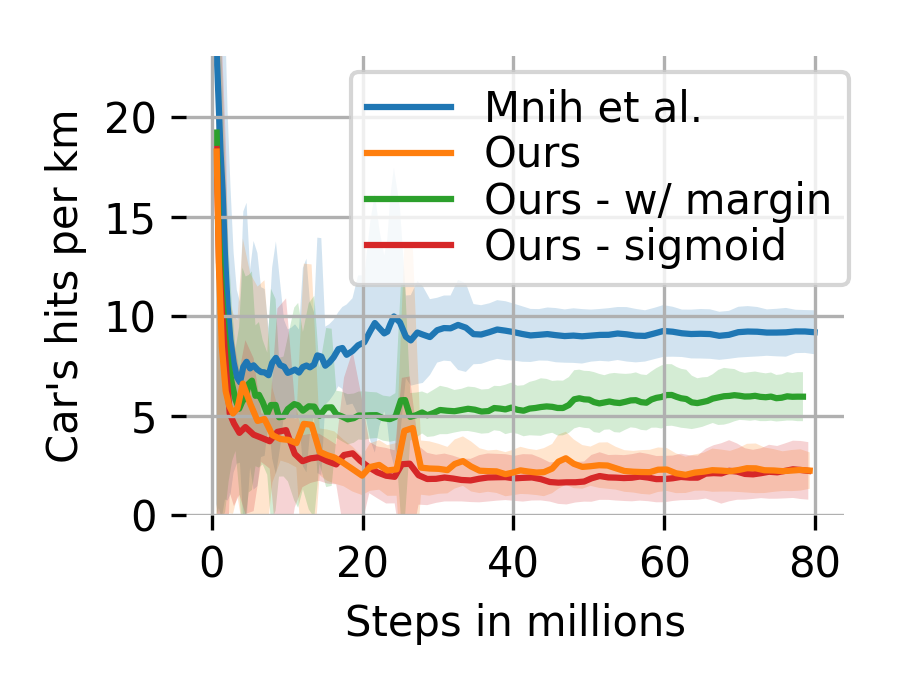}\label{fig:perfrewardrespawnReward}\hspace{-0.015\linewidth}\includegraphics[height=0.28\linewidth]{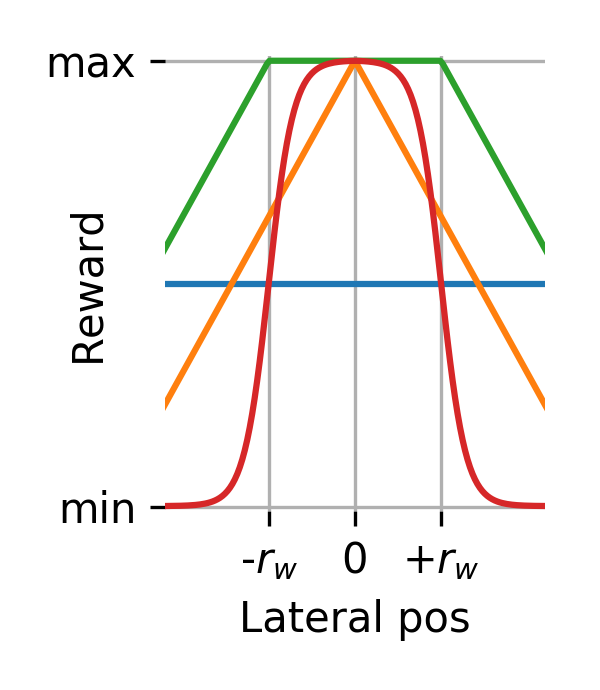}\label{fig:perfrewards}}\hspace{0.04\linewidth}\subfloat[Respawn strategy]{\includegraphics[height=0.28\linewidth]{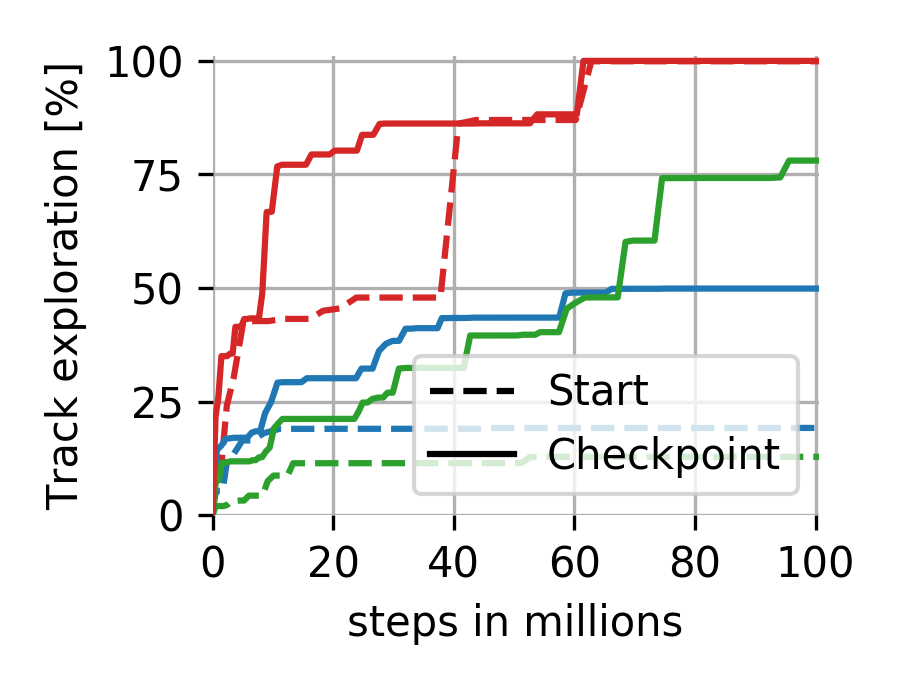}\label{fig:perfrewardrespawnRespawn}}
\caption{Evaluation of the performance of our strategies for reward (\ref{fig:perfrewardrespawnReward}) and respawn (\ref{fig:perfrewardrespawnRespawn}). In \ref{fig:perfrewardrespawnReward}, the reward used (orange) is compared to the reward from~\cite{Mnih2016AsynchronousLearning} (cyan). In \ref{fig:perfrewardrespawnRespawn}, our \textit{random checkpoint} strategy (plain line) and the \textit{start} (dash line) strategies for respawn are compared on the three tracks.}
\label{fig:perfrewardrespawn}
\end{figure}
We measure now the impact of the reward in terms driving style. In addition to compare reward \textit{Ours} (eq.~\ref{eq:rewardvcostd}) and \textit{Mnih et al.} (\cite{Mnih2016AsynchronousLearning}), we investigate the interest to account for the road width ($r_{w}$) in the reward and consequently propose two new rewards:
\begin{itemize}
	\item a reward penalizing distance only when the car is off the road lane: $R = v (\cos{\alpha} - max(d-0.5r_{w}, 0))$, named \textit{Ours w/ margin}.
	\item a smooth reward penalizing distance with a sigmoid: $R = v (\cos{\alpha} - \frac{1}{1+e^{-4(|d|-0.5r_{w})}})$, named \textit{Ours Sigmoid}.
\end{itemize}
Both of these rewards penalize more when the car leaves the road but the last one also avoids singularity. A visualization of the four rewards as a function of the car's lateral position is displayed in fig.~\ref{fig:perfrewardrespawnReward} right, assuming a constant speed for simplification.

Fig.~\ref{fig:perfrewardrespawnReward} left shows training performance as the number of car's hits per km. For a fair comparison the architecture of~\cite{Mnih2016AsynchronousLearning} is used and train on a single track. 
Our three rewards using the distance from track center lead to a significant drop in hits while converging faster than the reward \textit{Mnih et al.}, i.e.: Mnih et al. ($9.2$hits/km) versus Ours ($2.3$hits/km). Out of those three, the partly constant function \textit{Ours w/ margin} performs worst, probably because it is less suited for optimization via gradient descent. As could be expected, the fewer hits come at a cost of a slower overall speed: Mnih et al. ($106.3$km/h) compared to Ours ($91.4$km/h). To conclude Mnih et al. drives faster but with a much rougher style.



\subsubsection{Respawn strategy}
\label{sec:respawnstrat}
To evaluate our respawn strategy we compared to a similar network trained using the standard respawn at the start of the track. When the bots start at different positions the track completion is not a valid metric\footnote{E.g: a bot starting half way through the track could not reach more than 50\% completion. Hence, track completion would be a biased metric.}. Instead, we use the track exploration as a percentage of tracks length. In fig.~\ref{fig:perfrewardrespawnRespawn}, our random \textit{checkpoint} strategy (plain lines) exhibits a significantly better exploration than the usual \textit{start} strategy (dashed lines). For the easiest coast track (green), both strategies reach full exploration of the track but random checkpoint is faster. For complex snow (blue) and mountain (green) tracks, our strategy improves greatly the exploration by +32.20\% and +65.19\%, respectively.

The improvement due to our strategy is easily explained as it makes full use of the A3C decorrelation capability. Because it sees a wider variety of environments the network is forced generalize more. The fig.~\ref{fig:perfrewardrespawnRespawn} also shows that the bots tend to progress along the track in a non-linear fashion. The most logical explanation is that some track segments (e.g. sharp turns) require many attempts before bots succeed, leading to a training progress in fits and starts.
\\

To summarize, this comparative evaluation showed that our state encoder, respawn strategy and reward shaping improve greatly the end-to-end driving performance. Comparison of discrete versus continuous control was also conducted but not reported as they exhibit similar performances.

\section{Generalization}
\label{sec:expGeneralization}
\begin{figure}
	\subfloat[Speed limit]{\includegraphics[height=0.22\columnwidth]{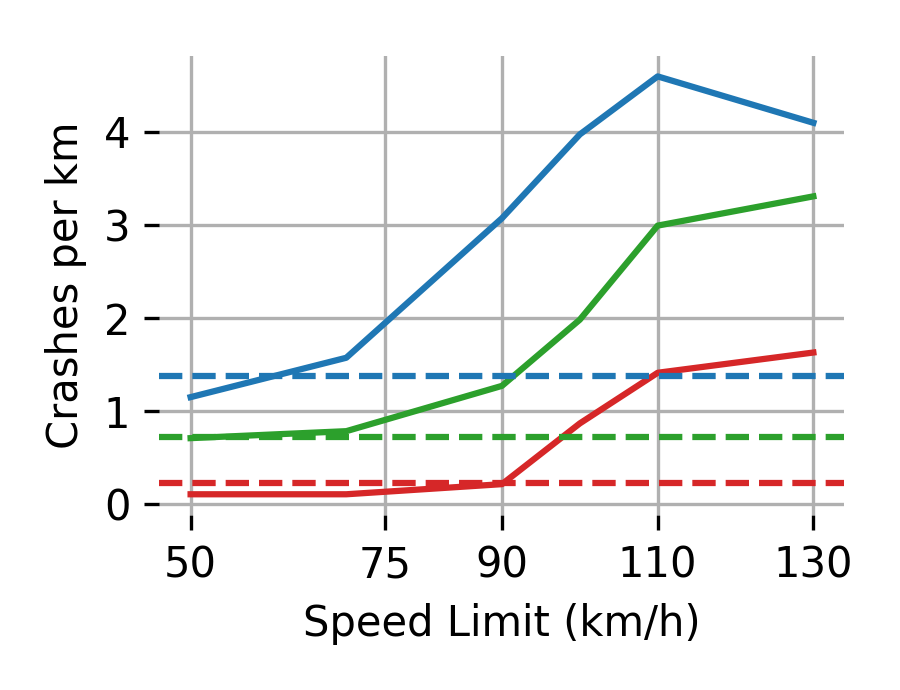}\label{fig:generalizationDrivingSpeedLimit}}\hspace{0.07\columnwidth}
	\subfloat[Prediction on real videos]{\includegraphics[height=0.204\columnwidth]{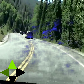}\hspace{0.01\columnwidth}\includegraphics[height=0.204\columnwidth]{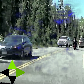}\hspace{0.01\columnwidth}\includegraphics[height=0.204\columnwidth]{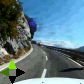}\label{fig:generalizationDrivingRealVideos}}
	\caption{\ref{fig:generalizationDrivingSpeedLimit} Influence of the speed limit on the number of car's hits for the three tracks and estimation of the hits with "real speed limit" (dash lines) calculated from the \textit{design speed}. Cf. sec.~\ref{sec:expGeneralizationSpeedLimit}. \ref{fig:generalizationDrivingRealVideos} Prediction of longitudinal and lateral commands on real videos. Note that the agent can handle situations never encountered (other road users, multi-lanes).}
	\label{fig:generalizationDriving}
\end{figure}

Because the WRC6 game is stochastic (animations, slight illumination changes, etc.) the performance reported on the training tracks already acknowledge some generalization. Still, we want to address these questions: Can the agent drive on unseen tracks? Can it drive respecting the speed limit? How does it perform on real images?

\subsubsection{Unseen tracks}
We tested the learned agent on tracks with different road layout which the agent could follow at high speeds, showing that the network incorporated general driving concepts rather than learned a track by heart. Qualitative performance is shown in the supplementary video.

\subsubsection{Racing VS Normal driving}
\label{sec:expGeneralizationSpeedLimit}
As the reward favors speed without direct penalizations of collisions the agent learned to go fast. While appropriate for racing games, this is dangerous for 'normal driving'. To evaluate how our algorithm could be transposed to normal driving with speed limits, we evaluate the influence of speed per track in fig. \ref{fig:generalizationDrivingSpeedLimit}. As one could expect, the number of crashes (collisions, off-roads) significantly reduces at lower speeds. Dashed lines in the figure show the performance in a real speed limit scenario. To estimate the real speed limit, we use the computed curvature and superelevation of each road segment to compute the ad-hoc \textit{design speed} as defined in the infrastructure standards~\cite{Policy2001}.

\subsubsection{Real videos}
Finally, we tested our agent on real videos (web footages, cropped and resize), and fig. \ref{fig:generalizationDrivingRealVideos} shows the guided back propagation and control output of a few frames. Although the results are partial as we cannot act on the video (i.e. control commands are never applied), the decision performance is surprisingly good for such a shallow network. In various environments the bot is capable of outputting the correct decision. To the best of our knowledge it is the first time a deep RL driving is shown working on real images and lets foresee that simulation based RL can be used as initialization strategy for decision making networks.

\section{Conclusion}
This paper introduced a framework and several learning strategies for end-to-end driving. The stochastic environment used is significantly more complex than existing ones and exhibits a variety of different physics/graphics with sharp road layouts. Compared to prior research, we learned the full control (lateral and longitudinal) including hand brake for drifts, in a completely self-supervised manner.
The comparative evaluation proved the importance of the learning strategy as our performance are above existing approaches despite the significantly more challenging task and the longer training tracks. Additionally, the bot shows generalization capacities in racing mode, and performance are even better with ad-hoc speed limits. Finally, experiments on real videos show our training can be transposed to camera images.

End-to-end driving is a challenging task and the research in that field is still at its earliest stage. These results participate to a step towards a real end-to-end driving platform.

{\small
\bibliographystyle{ieee}
\bibliography{bibliography}
}

\end{document}